\documentclass{article} 
\usepackage{iclr2026_conference,times}

\usepackage{amsmath,amsfonts,bm}

\def\eqref#1{equation~\ref{#1}}

\def\1{\bm{1}}

\DeclareMathAlphabet{\mathsfit}{\encodingdefault}{\sfdefault}{m}{sl}
\SetMathAlphabet{\mathsfit}{bold}{\encodingdefault}{\sfdefault}{bx}{n}

\usepackage{hyperref}
\usepackage{cleveref}

\usepackage{url}
\usepackage{microtype}
\usepackage{graphicx}
\usepackage{subcaption}
\usepackage{booktabs} 

\usepackage{enumitem}

\usepackage{amsmath}
\usepackage{amssymb}
\usepackage{mathtools}
\usepackage{amsthm}
\usepackage{multirow}
\usepackage{float}

\title{Understanding Empirical Unlearning with Combinatorial Interpretability}

\author{
Shingo Kodama\thanks{Equal contribution.} \\
Middlebury College
\And
Niv Cohen\footnotemark[1] \\
New York University
\And
Micah Adler \\
MIT
\And
Nir Shavit \\
MIT \& Red Hat
}

 \iclrfinalcopy 
\begin{document}

\maketitle

\begin{abstract}
While many recent methods aim to unlearn or remove knowledge from pretrained models, seemingly erased knowledge often persists and can be recovered in various ways. Because large foundation models are far from interpretable, understanding whether and how such knowledge persists remains a significant challenge.
To address this, we turn to the recently developed framework of combinatorial interpretability. This framework, designed for two-layer neural networks, enables direct inspection of the knowledge encoded in the model weights.
We reproduce baseline unlearning methods within the combinatorial interpretability setting and examine their behavior along two dimensions: (i) whether they truly remove knowledge of a target concept (the concept we wish to remove) or merely inhibit its expression while retaining the underlying information, and (ii) how easily the supposedly erased knowledge can be recovered through various fine-tuning operations.
Our results shed light within a fully interpretable setting on how knowledge can persist despite unlearning and when it might resurface.

\end{abstract}

\vspace{-3mm}

\section{Introduction}
Controlling the outputs of very large deep learning models is challenging, especially as these models are often trained on automatically collected data. As a result, they may emit during inference harmful, private, or otherwise undesired concepts \citep{carlini2021extracting,dong2025safeguarding,somepalli2023diffusion}. To address these issues, various recent unlearning methods have been proposed. Since unlearning methods are hard to model analytically and are usually evaluated empirically, strong mathematical guarantees are rare and mechanistic understanding of the erasure process remains limited.
Accordingly, these methods are vulnerable in both language and vision models \citep{lu2025concepts,zhangcatastrophic,lynch2024eight}. Here, we leverage the recently proposed framework of Combinatorial Interpretability to study the robustness of erasure methods in a more exact way.

A short overview of  existing methods for erasure and interpretability can be found in App.\ref{sec:related_works}

\vspace{-2mm}

\section{Background - Combinatorial Interpretability}

\vspace{-1mm}

We give below a short overview of the Combinatorial Interpretability framework, and refer the reader to \citet{adler2025towards} for an in-depth study.
A key limiting factor of current interpretability methods is representation entanglement or polysemanticity. Specifically, a single neuron’s activation often encodes multiple unrelated concepts. This issue becomes even more challenging as semantic concepts are not sharply defined and are not necessarily binary in nature.
To study such phenomena in a controlled setting, \citet{adler2025towards} explored simple two-layer neural networks trained on logical clauses. 
The simpler setting of 2-layer networks allows a direct inspection the knowledge in the network. For example, a network is trained on logical expressions such as DNFs:

\vspace{-2mm}
\begin{equation}
\begin{aligned}
&(x_6 \land x_{15} \land x_{11} \land x_{10})
\;\lor\;
(x_9 \land x_3 \land x_0 \land x_7)
\;\lor\;
(x_{12} \land x_5 \land x_2 \land x_4)
\;\lor\;
(x_{14} \land x_1 \land x_{13} \land x_8)
\end{aligned}
\label{eq:dnf-example}
\end{equation}
\vspace{-5mm}

\begin{figure}
    \centering
    \vspace{-5mm}\includegraphics[width=0.8\linewidth]{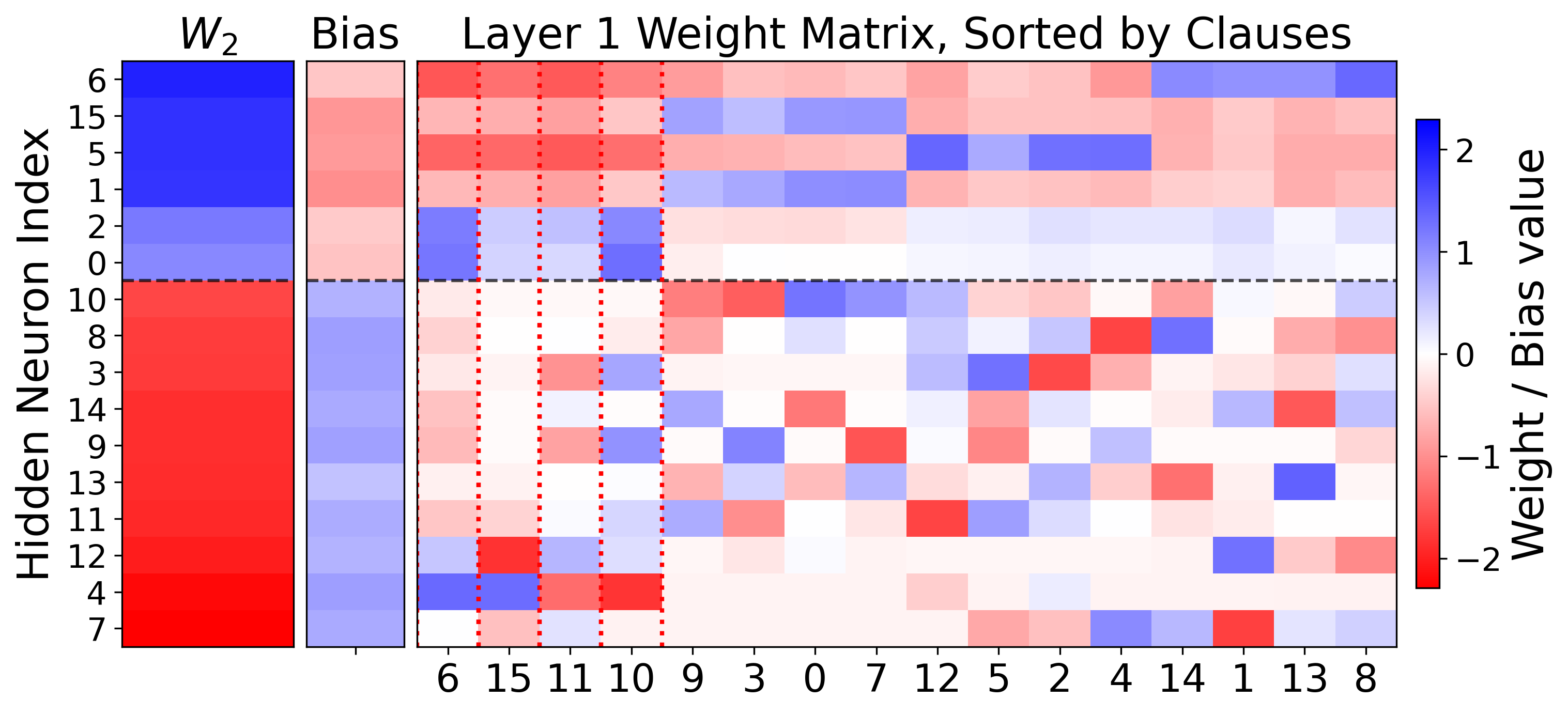}
    \caption{Weights of a 2-layer neural network trained on the Boolean formula in Eq.\ref{eq:dnf-example}. Row numbers indicate neurons, and column numbers in Layer 1 indicate input variables. ``Positive'' neurons (positive values in $W_2$) encode different clauses, with their corresponding weights being markedly positive. ``Negative'' neurons suppress spurious activations to inhibit misclassification.
    }
    \vspace{-5mm}
    \label{fig:comb_interp}
\end{figure}

\citet{adler2025towards} show that in the first layer of the network, the weights encode the clause structure of the formula (see Fig.\ref{fig:comb_interp}). Namely, to solve the logical clause identification task, the network uses two types of neurons in the first layer:
(i) The first type of neurons, termed ``positive neurons", encodes sets of weights aimed to identify specific clauses (possibly more than one clause per neuron), and impose a positive activation in the final classification neuron if any of the target clauses are detected. (ii) The second type, termed ``negative neurons'', inhibits spurious activations driven by variables not associated with any satisfied clause. Negative neurons act as a corrective mechanism: they detect these spurious activations and suppress them.
The second layer ideally activates the classification head if and only if at least one ``positive'' neuron detects a fulfilled clause.
Here, we treat each clause as representing a single concept, in analogy to concepts in an LLM or text-to-image model. This allows us to evaluate the removal of knowledge of certain concepts and the preservation of knowledge about others. 

\vspace{-2mm}
\section{Results}
\vspace{-2mm}

Here, we use Combinatorial Interpretability to examine three classical unlearning techniques. \textit{\textbf{Gradient Ascent} \citep{jang2023knowledge}}: This method inverses the training process. Instead of minimizing a loss function to learn a given dataset, the model maximizes the loss on the specific subset of data to be forgotten. 
\textit{\textbf{Task Vector} \citep{ilharco2022editing}}: A task vector is defined as the difference between the weights of a fine-tuned model, trained to strengthen its knowledge of a certain concept, and a pre-trained base model. To erase a specific capability, one applies the negative task vector to the base model, moving in parameter space in the opposite direction to that of learning.
\textit{
\textbf{Privacy Preserving Distillation (PPD)} \citep{papernot2016semi}}: This approach leverages the teacher-student framework to filter sensitive information. The student model is trained to mimic the general capabilities of a teacher model using only the non-senstive data.

\textbf{Datasets.}
We evaluate unlearning methods on synthetically generated DNF datasets (as in Eq.\ref{fig:comb_interp}). When variables are shared across different clauses, we refer to the logical expression as DNF-shared. 
Several hyperparameters control dataset generation, including the number of clauses, the size of each clause, the total number of input variables, and the number of clauses to be erased (targets). We vary them generating multiple reptetions of each described experiment.

\textbf{Recovery time.} 
We define recovery time as the number of fine-tuning steps required during recovery to reach an 80\% true positive rate (TPR) on the target clauses. Achieving a high TPR on the target clauses indicates that the model can once again correctly identify the erased clauses, and thus that the erased concept has been successfully revived.

Implementation details can be found in App.\ref{sec:implementation_details}.

\subsection{Persistence of knowledge and clause recovery time}

\begin{table}[t]
\centering
\small
\begin{tabular}{l l c c c c}
\toprule
Method & Data type & 80\% TPR @ (median) & 80\% Not Reached & Dist. Pos. & Dist Neg. \\
\midrule
Gradient Ascent  & Target      & 7 [6--9]     & 0\%   & 0.000 & 0.002 \\
Gradient Ascent  & Non Target  & 37 [28--48]  & 54\%  & 0.000 & 0.002 \\
Gradient Ascent  & All         & 9 [7--11]    & 0\%   & 0.000 & 0.002 \\
PPD & Target      & 17 [14--19]  & 0\%   & 1.013 & 0.989 \\
PPD & Non Target  & --           & 100\% & 1.013 & 0.989 \\
PPD & All         & --           & 100\% & 1.013 & 0.989 \\
Task Vector  & Target      & 10 [5--13]   & 0\%   & 0.004 & 0.017 \\
Task Vector  & Non Target  & 14 [8--22]   & 17\%  & 0.004 & 0.017 \\
Task Vector  & All         & 10 [6--14]   & 0\%   & 0.004 & 0.017 \\
\bottomrule
\end{tabular}

\caption{Recovery statistics of the target clause under different fine-tuning procedures (non-shared clauses). We report (i) the number of epochs it takes the model to achieve $80\%$ TPR on the target clause, (ii) the success rate of reaching this threshold within 60 epochs, and (iii) the cosine distance between the original clause structure and the post-erasure clause structure for each method.}
\label{tab:non-shared}
\end{table}

\vspace{-2mm}

In our first experiment, we train a Combinatorial Interpretability type model and examine when the supposedly erased concept can be recovered using different types of finetuning data.  Previous works have studied the time (number of fine-tuning epochs) required to recover a supposedly erased concept as a probe to determine how thoroughly this concept was erased \citep{golatkar2020eternal}. A recent work  shows that erased knowledge may re-emerge not only when fine-tuning on the target concept, but also when the model is fine-tuned on unrelated data \citep{george2025illusion}. We first evaluate this effect in a controlled setting using two-layer networks.

For each examined model, we apply each erasure technique to a degree that preserves the model’s ability to identify unrelated clauses (see App.\ref{sec:implementation_details}). Next, we finetune each erased model to potentially recover the erased clause in three ways: (i) using only samples that satisfy the \textit{target} clause (ii) using the entire initial dataset, containing \textit{all} clauses, target and non-target alike, and  (iii) using only unrelated (\textit{non-target}) clauses. 
In \Cref{tab:non-shared}, we report the results for formulae whose clauses are composed of distinct variables, and in \Cref{tab:shared} for formulae whose clauses share variables between them.

First, as expected, the erased target clause can be easily reintroduced across all erasure techniques when the model is fine-tuned on the target clause alone (\textit{target}). When fine-tuned on data from all clauses jointly, the concept is successfully reintroduced in models erased with Gradient Ascent or Task Vector, as these techniques largely preserve the clause encoding structure. Notably, in the Privacy Preserving Distillation based erasure, fine-tuning with all data struggles to re-introduce the concept even though the target concept is part of the finetuning data. This may happen as the distilled network learns a new encoding structure associated with the other clauses (see ~\Cref{fig:neuron_comp}) that does not easily accommodate additional clauses during finetuning. 

Surprisingly, we find that fine-tuning with Gradient Ascent– and Task Vector–based methods can recover the detectability of the target clause even when it is absent from the fine-tuning data. This suggests that the encoding of the network’s original knowledge remains attached to the edited weight structure. Erasing a concept without harming the detectability of other clauses leaves residual knowledge about the target clause in the model (see \textit{Dist. Pos.} and \textit{Dist. Neg.} for the average cosine distance between the original and erased models’ weights on positive and negative neurons; 0 indicates an identical match, and 1 indicates orthogonality). While we find that the fine-tuning process resurfaces existing knowledge this experiment does not explain how the recovery process occurs. We analyze this phenomenon further in the next section.

\vspace{-1mm}

\subsection{Decomposing recovery with unrelated fine-tuning.} Our findings suggest that erasure does not fully remove knowledge of the target concept. Instead, residual knowledge persists and can re-emerge during fine-tuning, even with unrelated clauses. 
We propose two hypotheses for how recovery with unrelated fine-tuning may occur:
\vspace{-1mm}

\begin{enumerate}
\item \textbf{Small perturbations may undo a shallow erasure.} In many cases, erasure leaves the model’s weights close to their pre-erasure values. Introducing only a small change that reduces the model’s ability to identify the target clause (see ~\Cref{fig:neuron_comp}). Therefore, fine-tuning on unrelated clauses may be sufficient to push the weights back into a regime where the target clause becomes identifiable again, even through a random-like change.

\item \textbf{Recovery may be driven by directed reversal.} The original model encodes the target clause using features that are partially shared with other clauses. When such representations are related, optimizing performance on unrelated clauses can still encourage weight updates that restore these shared features towards their initial encoding. Consequently, fine-tuning on unrelated clauses may  reverse the erasure of the target clause as well.
\end{enumerate}

To distinguish between these two cases, we run the following experiment:
For each neuron and clause we measure $d_f$, the weights difference between the model after concept erasure and the fine-tuned model (post erasure fine-tuning with unrelated data). We compute this difference for each neuron and for each clause resulting in a $clause\_size$-dimension vector. We then compare $d_f$ to $d_0$, the difference in the same $clause\_size$-dimension vector between the erased model and the original model before concept erasure. We decompose $d_f$ into vectors parallel and orthogonal to $d_0$, respectively $d_f^{\parallel}$ and $d_f^{\bot}$. We report 2 metrics:  (i) the magnitude of the recovery vector $\lVert d_f\rVert$, and (ii) the parallel fraction $\frac{\lVert d_f^{\parallel}\rVert}{\lVert d_f\rVert}$, which measures the fraction of the recovery magnitude that lies parallel to $d_0$. The update magnitude indicates which scenarios induces the largest changes. The parallel fraction helps interpret whether recovery fine-tuning primarily reverses the effects of erasure (high parallel fraction) or instead moves weights in a substantially different direction (low parallel fraction).

\begin{table}[t]
\centering
\small
\setlength{\tabcolsep}{5pt}

\begin{tabular}{llccccc}
\toprule
Fine-tune & Neuron sign & Clause type
& \multicolumn{2}{c}{Recovery Magnitude}
& \multicolumn{2}{c}{Parallel Fraction} \\
\cmidrule(lr){4-5} \cmidrule(lr){6-7}
 &  &  & DNF & DNF-shared & DNF & DNF-shared \\
\midrule
\multirow{4}{*}{Original Data}
& negative & control
& $0.14 \pm 0.08$ & $0.14 \pm 0.08$
& $0.97 \pm 0.06$ & $0.96 \pm 0.07$ \\
& negative & target
& $0.18 \pm 0.09$ & $0.17 \pm 0.09$
& $1.00 \pm 0.01$ & $0.99 \pm 0.01$ \\
& positive & control
& $0.13 \pm 0.08$ & $0.13 \pm 0.07$
& $0.94 \pm 0.14$ & $0.91 \pm 0.17$ \\
& positive & target
& $0.16 \pm 0.09$ & $0.16 \pm 0.09$
& $0.99 \pm 0.03$ & $0.97 \pm 0.06$ \\
\midrule
\multirow{4}{*}{Related}
& negative & control
& $0.18 \pm 0.08$ & $0.16 \pm 0.08$
& $0.99 \pm 0.02$ & $0.99 \pm 0.03$ \\
& negative & target
& $0.20 \pm 0.09$ & $0.19 \pm 0.09$
& $1.00 \pm 0.00$ & $1.00 \pm 0.01$ \\
& positive & control
& $0.11 \pm 0.12$ & $0.13 \pm 0.12$
& $1.00 \pm 0.02$ & $0.99 \pm 0.07$ \\
& positive & target
& $0.14 \pm 0.14$ & $0.16 \pm 0.14$
& $1.00 \pm 0.00$ & $0.99 \pm 0.06$ \\
\midrule
\multirow{4}{*}{Unrelated}
& negative & control
& $0.17 \pm 0.08$ & $0.16 \pm 0.08$
& $0.98 \pm 0.05$ & $0.97 \pm 0.06$ \\
& negative & target
& $0.19 \pm 0.08$ & $0.19 \pm 0.09$
& $0.99 \pm 0.03$ & $0.99 \pm 0.03$ \\
& positive & control
& $0.17 \pm 0.09$ & $0.17 \pm 0.08$
& $0.99 \pm 0.05$ & $0.96 \pm 0.11$ \\
& positive & target
& $0.18 \pm 0.10$ & $0.19 \pm 0.09$
& $1.00 \pm 0.01$ & $0.99 \pm 0.02$ \\
\bottomrule
\end{tabular}

\caption{
Magnitude of recovery vector and the fraction of it happening in the parallel direction (mean $\pm$ standard deviation) for DNF and DNF-shared models under Task Vector erasure, at epoch $60$ of recovery training (see ~\Cref{sec:additional_results} for similar results on Gradient Ascent based erasure). Results are averaged  across fine-tuning scenarios, neuron signs, and clause types, see App.\ref{sec:implementation_details}. 
}
\label{tab:recovery_mag_parallel_combined}

\vspace{-2mm}

\end{table}

Averaging over 13 scenarios of varying clause sizes, number of clauses, number of variables, and number of target clauses (see Table~\ref{tab:scenarios}), we obtain Table~\ref{tab:recovery_mag_parallel_combined}. We make two main observations: (1) There seems to be no statistically significant difference in the magnitude of the recovery vectors across neuron sign and clause type. Recovery works on all neurons, regardless of their role in the encoding of different clauses. (2) In all fine-tuning scenarios, there is consistent parallel fraction of close to $1$ throughout neuron signs and clause types. 
These findings suggest evidence for the second hypothesis for how recovery occurs. Since the model weights consistently move in the direction parallel to the weights movement during erasure, these results suggest that recovery is not well explained by random-like drift alone.

\section{Conclusion.}
The main findings of this paper are: (i) supposedly erased knowledge persists in model weights in a way that can be detected with Combinatorial Interpretability, (ii) this knowledge resurfaces when fine-tuning the erased model, even on non-target data, and (iii) recovery occurs through a tendency to restore the original (pre-erasure) network encoding. In future work we will use Combinatorial Interpretability to better understand when and why fine-tuning on non-target data restores the original encoding, and to assess the applicability of these findings to larger models used in practice.

\clearpage

\bibliography{iclr2026_conference}
\bibliographystyle{iclr2026_conference}

\clearpage

\appendix

\section{Implementation Details.}
\label{sec:implementation_details}

\textbf{Dataset generation}

Every generated assignment of variables contains at least one more active (True) variable than the size of the target clause, and the total number of active variables is restricted to a narrow range above this minimum. This ensures that positive examples are not simply constructed with True for all target clauses and False for all control clauses, requiring the model to identify the target clause among noisy variables.

For DNF-shared, we have additional hyperparameters. I.e., as the number of variables overlapping between target and non-target (control) clauses, and the number of control clauses that share variables with the target clauses.

\textbf{Unlearning techniques}

Each unlearning method involves multiple hyperparameters. To tune these, we performed a grid search on a fixed setting consisting of four clauses of size four, 16 input variables, one target clause, and a DNF structure. Hyperparameters were selected to maximize the TPR for control clauses while enforcing a TPR of 0 on the target clause. This ensures that the target concept has been fully erased, while also preserving the mdoel's ability to identify control clauses. Unless otherwise stated, we use the following hyperparameter configurations throughout the paper.
For Gradient Ascent, we use a learning rate of $0.005$ and perform $70$ finetuning steps. For Task Vector, we use a learning rate of $0.005$, $50$ finetuning steps, and a scaling factor of $1.0$, which controls the magnitude of the negative task vector applied to the base model. For Privacy Preserving Distillation, several distillation strategies are possible, including hard labels, temperature-scaled probabilities, and raw logits. In this work, we adopt temperature-scaled probability distillation with a learning rate of $0.01$, $100$ fine-tuning steps, and a temperature of $1.5$.

\textbf{TPR and FPR.}

For both target and control clauses, we measure the true positive rate (TPR) and false positive rate (FPR). For each assigment type (target clause or other clauses), we generate $2,000$ samples of assignments that satisfy only that clause, as well as $2,000$ samples of assignments that satisfy no clauses. Using these samples, we compute the TPR and FPR for each individual clause. We then average these values across all target clauses and across all control clauses to obtain the overall TPR and FPR for target and control clauses, respectively.

\textbf{Recovery fine-tuning}

For concept recovery, we fine-tune the model for 60 steps using a learning rate of 0.0002. Fine-tuning is performed on 10,000 samples, drawn from either unrelated clauses, related clauses, or all clauses. Model performance is evaluated using 2,000 positive samples (assignments that satisfy the given clause) and 2,000 negative samples (assignments that satisfy no clause), for each assignment type.

\begin{table}[H]
\centering
\small
\setlength{\tabcolsep}{6pt}
\begin{tabular}{c c c c}
\toprule
Scenario & num\_clauses & clause\_size & clauses\_to\_erase \\
\midrule
1  & 4 & 4 & [0] \\
2  & 4 & 5 & [0] \\
3  & 5 & 5 & [0] \\
4  & 5 & 6 & [0] \\
5  & 6 & 6 & [0] \\
6  & 6 & 6 & [0,1] \\
7  & 7 & 7 & [0] \\
8  & 7 & 7 & [0,1] \\
9  & 8 & 8 & [0] \\
10 & 8 & 8 & [0,1] \\
11 & 5 & 4 & [0] \\
12 & 5 & 4 & [0,1] \\
13 & 6 & 4 & [0] \\
\bottomrule
\end{tabular}
    \caption{The 13 experimental scenarios used throughout the paper. The number variables is calculated as $num\_clauses\times clause\_size$.}
\label{tab:scenarios}
\end{table}

\section{Related Works}
\label{sec:related_works}

\textbf{Unlearning and Concept Erasure.} Machine unlearning refers to techniques that remove or forget specific training data or concepts from a trained model. As exact methods show little improvement over full retraining, a lot of focus is directed to removing specific concepts from a model’s representations rather than forgetting particular data instance \citep{warnecke2021machine}. Concept erasure, or concept removal, focuses on removing attributes from a model’s representations or behavior, rather than forgetting particular data instances. In NLP, recent work removes attributes by training an auxiliary classifier and optimizing the model such that this classifier can no longer detect the attribute \citep{ravfogel2022linear,iskander2023shielded}. Other works focus on evading or removing specific pieces of knowledge\citep{gur2025precise,lynch2024eight}.

In vision generative models, many methods have been proposed to remove concepts or prevent them from being generated \citep{gandikota2023erasing,schramowski2023safe,gong2024reliable}. Yet such approaches are often vulnerable as various intervention may resrface the supposly erased knoweldge \cite{lu2025concepts,lynch2024eight,george2025illusion}.

\textbf{Mechanistic interpretability.}
Mechanistic interpretability research seeks to reverse-engineer neural networks to identify human-understandable component that correspond to meaningful functions \citep{radford2018learning,clark2019does,tenney2019bert}. Some approaches use causal tracing method to locate which internal activations carry a given factual association \citep{meng2022locating}, while others look for circuits in the network weights. Such circuits are sets of neurons and attention heads that implement an interpretable function \cite{olsson2022context}. 

\textbf{Polysemanticity.} \citet{elhage2022toy}  showed that networks often store many unrelated features in the same neuron, a phenomenon described as polysemanticity or superposition. This effect may be encouraged as neural layers have limited dimension. \citet{adler2024complexity} explored the theoretical limits of optimal encoding in superposition. These ideas were later extended to interpretability studies of simple two-layer networks \citep{adler2025towards}.

\section{Additional Results}
\label{sec:additional_results}

\textbf{Comparison of Erased Model Weights}

\begin{figure}[H]
    \centering
    \includegraphics[width=0.8\linewidth]{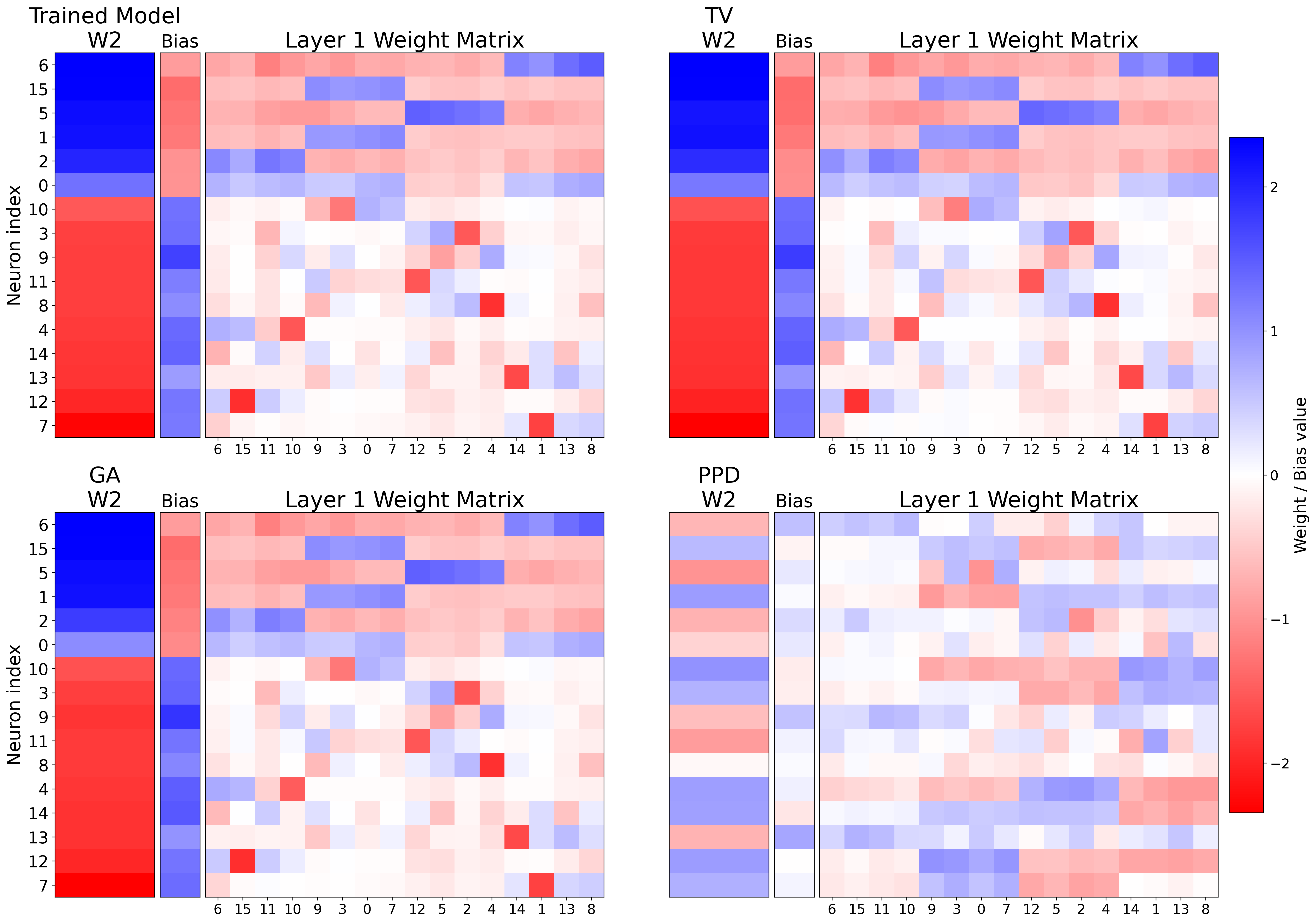}
    \caption{Model weights of original trained model and models after concept erasure. Models erased using Task Vectors or Gradient Ascent remain very close to the original model, whereas PPD yields a model with substantially different internal encodings because it is trained as a student network using random initialization.}
    \label{fig:neuron_comp}
\end{figure}

\textbf{Recovery with different types of data for clauses with shared variables.}

\begin{table}[H]
\centering
\small
\begin{tabular}{l l c c c c}
\toprule
Method & Data type & 80\% TPR @ (median) & 80\% Not Reached & Disat. Pos. & Dist. Neg. \\
\midrule
Gradient Ascent  & Target      & 6 [5--8]     & 0\%   & 0.000 & 0.001 \\
Gradient Ascent  & Non Target  & 25 [18--27]  & 17\%  & 0.000 & 0.001 \\
Gradient Ascent  & All         & 8 [7--9]     & 0\%   & 0.000 & 0.001 \\
PPD & Target      & 14 [13--16]  & 0\%   & 0.971 & 1.003 \\
PPD & Non Target  & 59 [59--59]  & 94\%  & 0.971 & 1.003 \\
PPD & All         & --           & 100\% & 0.971 & 1.003 \\
Task Vector  & Target      & 5 [5--12]    & 0\%   & 0.003 & 0.007 \\
Task Vector  & Non Target  & 6 [6--17]    & 0\%   & 0.003 & 0.007 \\
Task Vector  & All         & 6 [6--13]    & 0\%   & 0.003 & 0.007 \\
\bottomrule
\end{tabular}

\caption{Recovery statistics of the target clause under different fine-tuning procedures (shared clauses). We report (i) the number of epochs it takes the model to achieve $80\%$ TPR on the target clause, (ii) the success rate of reaching this threshold within $60$ epochs, and (iii) the cosine distance between the original clause structure and the post-erasure clause structure for each method.}
\label{tab:shared}
\end{table}

\textbf{Recovery Decomposition for Gradient Ascent}

\begin{table}[H]
\centering
\small
\setlength{\tabcolsep}{5pt}

\begin{tabular}{llccccc}
\toprule
Fine-tune & Neuron sign & Clause type
& \multicolumn{2}{c}{Recovery Magnitude}
& \multicolumn{2}{c}{Parallel Fraction} \\
\cmidrule(lr){4-5} \cmidrule(lr){6-7}
 &  &  & DNF & DNF-shared & DNF & DNF-shared \\
\midrule
\multirow{4}{*}{Original Data}
& negative & control
& $0.05 \pm 0.02$ & $0.06 \pm 0.02$
& $0.73 \pm 0.27$ & $0.75 \pm 0.22$ \\
& negative & target
& $0.14 \pm 0.03$ & $0.13 \pm 0.03$
& $0.98 \pm 0.07$ & $0.98 \pm 0.05$ \\
& positive & control
& $0.07 \pm 0.04$ & $0.08 \pm 0.03$
& $0.75 \pm 0.27$ & $0.71 \pm 0.26$ \\
& positive & target
& $0.12 \pm 0.06$ & $0.12 \pm 0.06$
& $0.90 \pm 0.19$ & $0.85 \pm 0.28$ \\
\midrule
\multirow{4}{*}{Related}
& negative & control
& $0.13 \pm 0.03$ & $0.12 \pm 0.03$
& $0.99 \pm 0.03$ & $0.81 \pm 0.18$ \\
& negative & target
& $0.15 \pm 0.03$ & $0.14 \pm 0.04$
& $1.00 \pm 0.00$ & $1.00 \pm 0.00$ \\
& positive & control
& $0.12 \pm 0.14$ & $0.11 \pm 0.11$
& $0.99 \pm 0.03$ & $0.82 \pm 0.21$ \\
& positive & target
& $0.16 \pm 0.18$ & $0.14 \pm 0.11$
& $1.00 \pm 0.00$ & $1.00 \pm 0.01$ \\
\midrule
\multirow{4}{*}{Unrelated}
& negative & control
& $0.09 \pm 0.02$ & $0.09 \pm 0.02$
& $0.96 \pm 0.06$ & $0.77 \pm 0.19$ \\
& negative & target
& $0.10 \pm 0.02$ & $0.10 \pm 0.02$
& $0.98 \pm 0.03$ & $0.98 \pm 0.03$ \\
& positive & control
& $0.10 \pm 0.04$ & $0.11 \pm 0.05$
& $0.99 \pm 0.02$ & $0.83 \pm 0.19$ \\
& positive & target
& $0.11 \pm 0.03$ & $0.12 \pm 0.04$
& $0.99 \pm 0.04$ & $0.98 \pm 0.02$ \\
\bottomrule
\end{tabular}
\caption{
Magnitude of recovery vector and the fraction of it happening in the parallel direction (mean $\pm$ standard deviation) for DNF and DNF-shared models under Gradient Ascent erasure, at epoch $60$ of recovery training. Results are averaged  across fine-tuning scenarios, neuron signs, and clause types, see App.\ref{sec:implementation_details}. 
}
\label{tab:recovery_mag_parallel_combined_ga}

\vspace{-2mm}
\end{table}

We see similar results for Gradient Ascent erasure, though the parallel fraction is not as consistently close to $1$ as in Task Vector erasure.

\end{document}